\documentclass[11pt]{article}

\usepackage[margin=1in]{geometry}
\usepackage{array}
\usepackage{booktabs}
\usepackage{caption}
\usepackage{graphicx}
\usepackage{float}
\usepackage{placeins}
\usepackage{xurl}
\usepackage[colorlinks=true,linkcolor=blue,citecolor=blue,urlcolor=blue]{hyperref}
\usepackage{microtype}
\usepackage[round]{natbib}
\usepackage{tabularx}
\usepackage{xcolor}
\usepackage{tikz}
\usetikzlibrary{arrows.meta,positioning,calc}
\definecolor{ledgerblue}{HTML}{2F6F9F}
\definecolor{ledgergreen}{HTML}{4D8B57}
\definecolor{ledgerorange}{HTML}{C77C2C}
\definecolor{ledgergray}{HTML}{6B7280}
\definecolor{ledgerred}{HTML}{B64A4A}
\definecolor{ledgerpurple}{HTML}{7A5AA6}

\title{Evidence-Ledger Adjudication for Claim-Evidence Traceability}
\author{
Gengyu Chen\\
Carnegie Mellon University
\and
Yongjie Yu\\
Carnegie Mellon University
\and
Weiling Wang\\
Syracuse University
}
\date{June 26, 2026}

\begin{document}
\maketitle

\begin{abstract}
AI agents can draft claims faster than authors can check whether the cited or retrieved evidence supports them. We study evidence-ledger adjudication: a claim-evidence traceability workflow that pairs each claim with an evidence packet, assigns a support relation, and routes unsupported, contradicted, or mixed-evidence claims back to the author. The empirical core is a 2,335-row blind benchmark built from independent external labels in AVeriTeC, CLIMATE-FEVER, and SciFact. Gold relations and source evidence labels are hidden during prediction and joined only for scoring. On this benchmark, the agent evidence-ledger condition achieves 0.676 relation accuracy and 0.601 macro-F1, compared with 0.383 accuracy and 0.303 macro-F1 for the best non-agent baseline. It also routes 1270/1435 claims whose gold labels indicate contradiction, missing evidence, or mixed evidence, while routing 295/900 supported claims. These results show that evidence-ledger adjudication can turn heterogeneous evidence packets into an auditable traceability layer for AI-assisted writing.
\end{abstract}

\section{Introduction}

AI-assisted writing changes the bottleneck in research communication. A language model can produce literature summaries, draft claims, and citation-shaped prose quickly, but authors still need to decide whether the available evidence supports each claim. This problem is practical rather than cosmetic: a fluent sentence can be unsupported, contradicted, or only partly established by the evidence it cites. Reviewers often encounter these problems late, after they have already affected the manuscript's clarity and credibility.

This paper studies evidence-ledger adjudication for claim-evidence traceability in AI-assisted writing. The workflow records a claim, its evidence packet, a predicted support relation, an author-review route flag, confidence, and a short rationale. The output is meant to be read by authors while claims are still being shaped: supported claims can stay in the draft, while unsupported, contradicted, and mixed-evidence claims are routed for revision, additional evidence search, or removal.

The central empirical question is whether this author-side workflow transfers across independently labeled claim-verification sources. We therefore evaluate against external labels from AVeriTeC \citep{Schlichtkrull2023AVeriTeC}, CLIMATE-FEVER \citep{Diggelmann2020ClimateFever}, and SciFact \citep{Wadden2020Fact}. The resulting blind benchmark contains 2,335 test rows spanning real-world fact-checks, climate claims, and scientific claims. It uses four normalized relations: supports, contradicts, missing evidence, and mixed evidence.

We make three contributions.

\begin{enumerate}
  \item We define an evidence-ledger adjudication workflow that turns claim/evidence packets into explicit support relations and author-review routes.
  \item We build a broad external evaluation protocol from AVeriTeC, CLIMATE-FEVER, and SciFact, with gold relations and source evidence labels hidden from the prediction step.
  \item We show that the agent evidence-ledger condition substantially improves over always-supported, lexical, and TF-IDF logistic baselines on relation accuracy, macro-F1, and routing recall for claims that need author attention.
\end{enumerate}

\section{Related Work}

\subsection{External claim-verification benchmarks}

SciFact introduced scientific claim verification against cited literature evidence \citep{Wadden2020Fact}. AVeriTeC extends real-world claim verification with question-answer evidence and verdict labels from professional fact-checking contexts \citep{Schlichtkrull2023AVeriTeC}. CLIMATE-FEVER focuses on real-world climate claims and human-annotated evidence sentences \citep{Diggelmann2020ClimateFever}. These datasets are valuable for this paper because their labels were created independently of the present workflow.

\subsection{Citation and manuscript checking}

SemanticCite verifies citation statements with evidence-based reasoning over source text \citep{Haan2025SemanticCite}. \texttt{sciwrite-lint} frames manuscript verification as infrastructure for checking references, consistency, figures, and text \citep{Samsonau2026SciwriteLint}. FactReview studies evidence-grounded peer review with optional execution checks \citep{Yue2026FactReview}, and Peerispect verifies claims in peer reviews against manuscripts \citep{Ghorbanpour2026Peerispect}. Our workflow is positioned during drafting: it asks the authoring agent to expose support relations while authors can still revise the claim, evidence, or wording.

\subsection{Agent evaluation through artifacts}

Agent benchmarks increasingly evaluate behavior through traces, tools, states, and artifacts rather than final text alone. AgentBench evaluates agents across environments \citep{Liu2023AgentBench}; SWE-bench ties success to project state \citep{Jimenez2023bench}; WebArena evaluates agents in realistic web environments \citep{Zhou2023WebArena}; ToolLLM studies tool use over API collections \citep{Qin2023ToolLLM}; and AgentBeats emphasizes open, standardized, reproducible agent assessment \citep{Liu2026AgentBeats}. Evidence-ledger adjudication applies the same artifact-grounded idea to writing: the claim, evidence packet, relation, route flag, and rationale become reviewable intermediate artifacts.

\section{Evidence-Ledger Adjudication}

\subsection{Workflow}

\begin{figure}[H]
\centering
\begin{tikzpicture}[
  node distance=0.48cm,
  box/.style={draw, rounded corners=2pt, align=center, minimum height=1.00cm, text width=2.35cm, font=\small},
  arrow/.style={-{Latex[length=2.2mm]}, thick, ledgergray}
]
\node[box, fill=ledgerblue!10, draw=ledgerblue] (claim) {claim};
\node[box, fill=ledgergreen!10, draw=ledgergreen, right=of claim] (evidence) {evidence\\packet};
\node[box, fill=ledgerpurple!10, draw=ledgerpurple, right=of evidence] (relation) {support\\relation};
\node[box, fill=ledgerorange!10, draw=ledgerorange, right=of relation] (route) {author\\route};
\node[box, fill=gray!12, draw=ledgergray, right=of route] (action) {revise, search,\\or keep};
\draw[arrow] (claim) -- (evidence);
\draw[arrow] (evidence) -- (relation);
\draw[arrow] (relation) -- (route);
\draw[arrow] (route) -- (action);
\draw[arrow, dashed] (action.south) .. controls +(0,-0.82) and +(0,-0.82) .. node[below, font=\scriptsize, align=center] {claim revision, evidence search,\\or author confirmation} (claim.south);
\end{tikzpicture}
\caption{Evidence-ledger adjudication turns a claim and its evidence packet into an explicit relation and author-review route.}
\label{fig:workflow}
\end{figure}
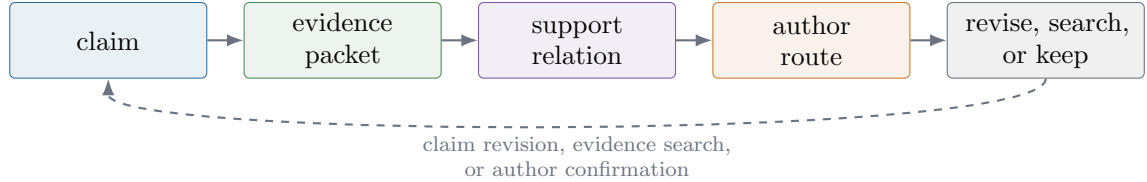

Figure~\ref{fig:workflow} shows the proposed workflow. A claim enters the ledger with its evidence packet. The adjudicator reads the packet, assigns one relation, and emits an author-review route. The relation is the evaluation target; the route is the author-facing action.

\subsection{Relations and route rule}

The adjudicator chooses exactly one relation. \emph{Supports} means the evidence establishes the claim. \emph{Contradicts} means the evidence conflicts with the claim. \emph{Missing evidence} means the packet does not establish the claim. \emph{Mixed evidence} means the packet supports part of the claim but challenges another part, or contains competing evidence that should be reviewed together.

The author-review route is enabled for contradiction, missing evidence, and mixed evidence, and disabled for supported claims. This route rule converts relation prediction into a writing action: claims that need attention are sent back to the authoring loop, while supported claims are not.

\section{External Benchmark Protocol}

\subsection{Sources and normalization}

\begin{figure}[H]
\centering
\includegraphics[width=0.92\textwidth]{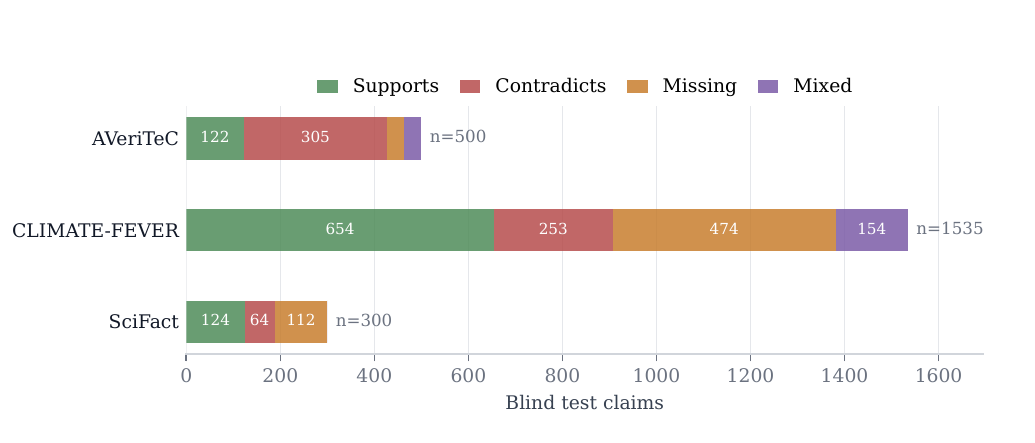}
\caption{Broad external benchmark composition by source and normalized relation label. The blind test packet contains 2,335 rows.}
\label{fig:composition}
\end{figure}

The benchmark combines three independently labeled external sources. AVeriTeC contributes 500 development-set rows with question-answer evidence packets. CLIMATE-FEVER contributes 1,535 climate-claim rows with released evidence sentences. SciFact contributes 300 development-set rows with cited scientific abstracts. Figure~\ref{fig:composition} shows the source and label composition.

Source labels are normalized into the four evidence-ledger relations. AVeriTeC's supported, refuted, and not-enough-evidence labels map to supports, contradicts, and missing evidence; its cherry-picking verdict maps to mixed evidence. CLIMATE-FEVER's supports, refutes, not-enough-information, and disputed labels map to the same four relations. SciFact support and contradiction labels map directly, and claims without supporting cited evidence map to missing evidence.

\subsection{Blind packet}

\begin{figure}[H]
\centering
\includegraphics[width=0.90\textwidth]{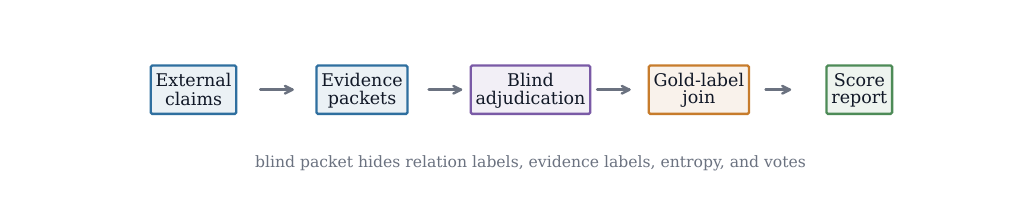}
\caption{Blind protocol. Gold relations and source evidence labels are hidden during adjudication and joined only during scoring.}
\label{fig:protocol}
\end{figure}

Figure~\ref{fig:protocol} summarizes the blind protocol. Prediction inputs contain claim text, source name, broad field metadata, evidence documents, allowed relations, and the route rule. They do not contain gold relations, source labels, evidence labels, entropy values, or evidence-vote counts. For CLIMATE-FEVER, released evidence annotations are used only to select which source-provided evidence sentences enter the packet; the annotations themselves are removed before prediction.

The broad test packet has 2,335 rows: 900 supports, 622 contradicts, 621 missing-evidence cases, and 192 mixed-evidence cases. The non-agent baselines use 3,877 labeled tuning rows from the available external training data. The agent condition is evaluated on the blind test packet in 47 batches, with row-count, claim-order, prompt-hash, and prediction-hash validation recorded after the run. Because the source benchmarks are public, the blind design controls prediction-time access to labels and packet metadata rather than model pretraining exposure.

\begin{table}[H]
\centering
\caption{External benchmark protocol.}
\label{tab:protocol}
\small
\begin{tabularx}{\textwidth}{>{\raggedright\arraybackslash}p{0.25\textwidth}>{\raggedright\arraybackslash}X}
\toprule
Design item & Protocol choice \\
\midrule
External labels & AVeriTeC, CLIMATE-FEVER, and SciFact labels created independently of this workflow. \\
Blind test packet & 2,335 claim/evidence rows with gold relations and source evidence labels removed. \\
Tuning data & 3,877 labeled external rows used only for non-agent baseline tuning. \\
Agent condition & 47 blind batches; predictions validated for row count and claim-id order before scoring. \\
Relations & Supports, contradicts, missing evidence, and mixed evidence. \\
Primary metrics & Relation accuracy, macro-F1, review-needed recall, and supported-claim routing rate. \\
\bottomrule
\end{tabularx}
\end{table}

\subsection{Conditions}

We compare four conditions. The always-supported baseline predicts that every claim is supported. The lexical-ledger baseline uses train-tuned lexical overlap and contradiction cues. The TF-IDF logistic baseline trains a balanced multiclass classifier on the tuning rows. The agent evidence-ledger condition reads each blind packet and emits a relation, route flag, confidence, and short rationale.

These conditions answer distinct reviewer questions. Always-supported measures the value of doing any attention routing. The lexical baseline tests whether shallow traceability rules are enough. The TF-IDF baseline tests a lightweight supervised classifier over claim/evidence text. The agent condition tests the final evidence-ledger workflow.

\section{Results}

\subsection{Overall benchmark performance}

Table~\ref{tab:results} reports the main result. The agent evidence-ledger condition reaches 1579/2335 relation accuracy (0.676) and macro-F1 0.601. The best non-agent baseline reaches 895/2335 relation accuracy (0.383) and macro-F1 0.303. The agent condition also routes 1270/1435 claims labeled contradiction, missing evidence, or mixed evidence, compared with 1038/1435 for the best non-agent baseline on that metric.

\begin{table}[H]
\centering
\caption{Broad external benchmark results. Review-needed recall is recall over contradiction, missing-evidence, and mixed-evidence labels. Supported routed is lower when supported claims are left alone.}
\label{tab:results}
\small
\begin{tabular}{lrrrr}
\toprule
Condition & Relation accuracy & Macro-F1 & Review recall & Supported routed \\
\midrule
Always supported & 900 / 2335 (0.385) & 0.139 & 0 / 1435 (0.000) & 0 / 900 (0.000) \\
Lexical ledger & 856 / 2335 (0.367) & 0.291 & 941 / 1435 (0.656) & 417 / 900 (0.463) \\
TF-IDF logistic & 895 / 2335 (0.383) & 0.303 & 1038 / 1435 (0.723) & 547 / 900 (0.608) \\
Agent evidence ledger & 1579 / 2335 (0.676) & 0.601 & 1270 / 1435 (0.885) & 295 / 900 (0.328) \\
\bottomrule
\end{tabular}
\end{table}

\begin{figure}[H]
\centering
\includegraphics[width=0.92\textwidth]{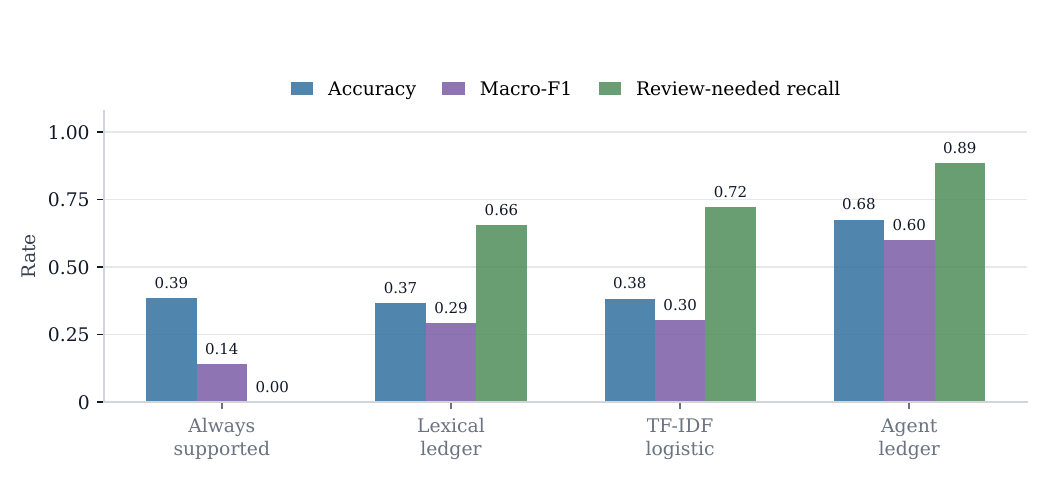}
\caption{Main benchmark comparison. The agent evidence-ledger condition improves over the non-agent baselines on relation accuracy, macro-F1, and review-needed recall.}
\label{fig:main-results}
\end{figure}

\begin{figure}[H]
\centering
\includegraphics[width=0.86\textwidth]{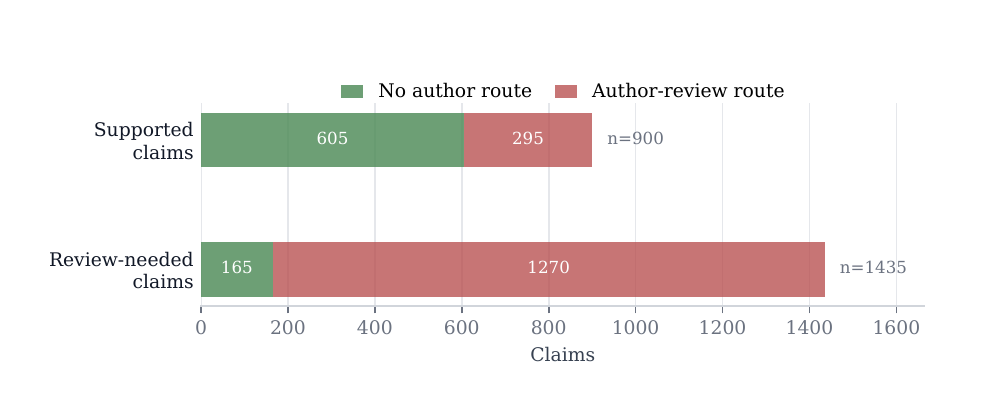}
\caption{Author-review routing produced by the agent evidence-ledger condition. The route sends unsupported, contradicted, and mixed-evidence claims back to the authoring loop.}
\label{fig:routing}
\end{figure}

Figure~\ref{fig:routing} shows the author-review workload implied by the route rule. The agent routes most claims that need attention while keeping most supported claims out of the author-review queue.

\subsection{Source and label behavior}

Table~\ref{tab:source-results} reports source-level results. The agent condition performs strongly on AVeriTeC and SciFact, and CLIMATE-FEVER contributes the largest source slice in the benchmark.

\begin{table}[H]
\centering
\caption{Agent evidence-ledger results by external source.}
\label{tab:source-results}
\small
\begin{tabular}{lrrrr}
\toprule
Source & Claims & Relation accuracy & Macro-F1 & Review recall \\
\midrule
AVeriTeC & 500 & 422 / 500 (0.844) & 0.714 & 364 / 378 (0.963) \\
CLIMATE-FEVER & 1535 & 901 / 1535 (0.587) & 0.527 & 748 / 881 (0.849) \\
SciFact & 300 & 256 / 300 (0.853) & 0.643 & 158 / 176 (0.898) \\
\bottomrule
\end{tabular}
\end{table}

\begin{figure}[H]
\centering
\includegraphics[width=0.82\textwidth]{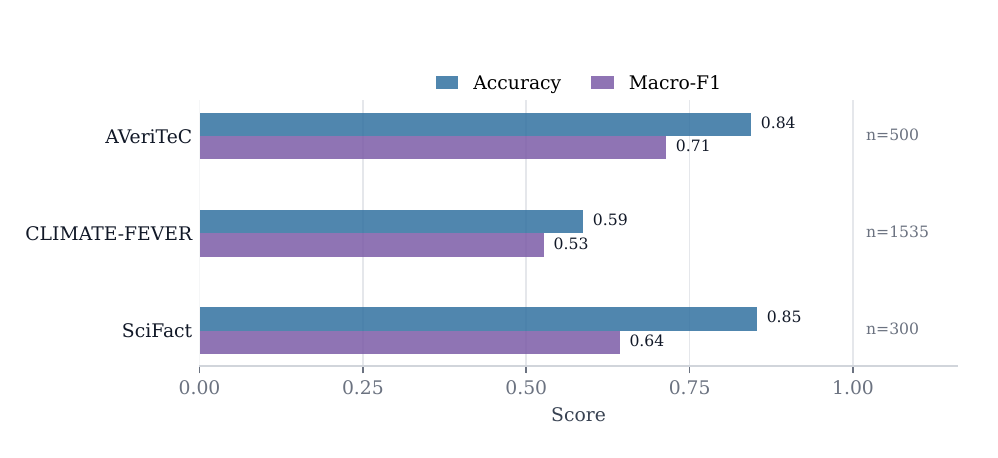}
\caption{Source-level performance. AVeriTeC and SciFact are the strongest slices; CLIMATE-FEVER is the largest source slice.}
\label{fig:source-performance}
\end{figure}

Table~\ref{tab:label-results} reports label-level precision, recall, and F1. Supports and contradicts are the strongest labels, missing evidence is moderate, and mixed evidence is reported as a smaller diagnostic relation.

\begin{table}[H]
\centering
\caption{Agent evidence-ledger performance by normalized external label.}
\label{tab:label-results}
\small
\begin{tabular}{lrrrrr}
\toprule
Gold relation & Claims & Correct & Precision & Recall & F1 \\
\midrule
Supports & 900 & 605 & 0.786 & 0.672 & 0.725 \\
Contradicts & 622 & 502 & 0.778 & 0.807 & 0.792 \\
Missing & 621 & 428 & 0.530 & 0.689 & 0.599 \\
Mixed evidence & 192 & 44 & 0.389 & 0.229 & 0.289 \\
\bottomrule
\end{tabular}
\end{table}

\begin{figure}[H]
\centering
\includegraphics[width=0.86\textwidth]{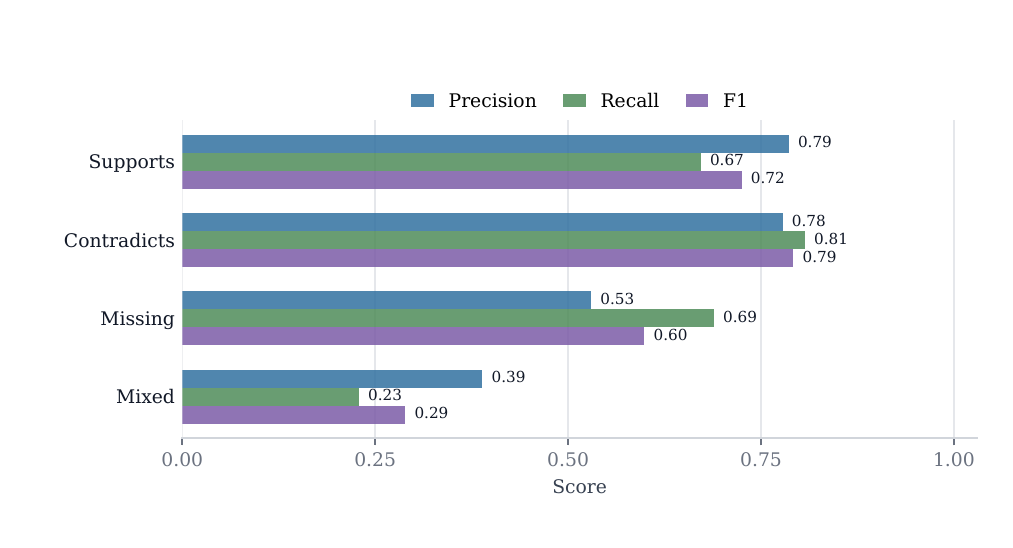}
\caption{Label-level precision, recall, and F1 for the agent evidence-ledger condition.}
\label{fig:label-performance}
\end{figure}

\begin{figure}[H]
\centering
\includegraphics[width=0.68\textwidth]{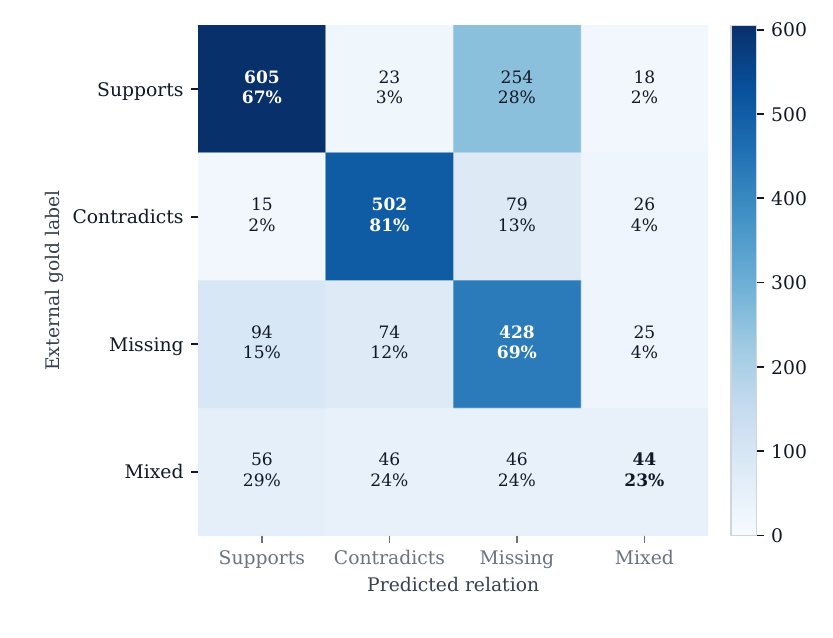}
\caption{Agent confusion matrix on the 2,335-row broad external benchmark. Rows are external gold labels and columns are predicted relations.}
\label{fig:confusion}
\end{figure}
\FloatBarrier

\subsection{Positioning against adjacent systems}

Table~\ref{tab:close-tools} positions the workflow against adjacent systems. The table is qualitative because these systems target different parts of the research and review pipeline. The empirical comparison in this paper is against the baselines in Table~\ref{tab:results}; the adjacent-system table clarifies task scope.

\begin{table}[H]
\centering
\caption{Qualitative task positioning. ``Direct'' means the evidence class is a primary task for the system; ``partial'' means related support exists but is not the main task.}
\label{tab:close-tools}
\scriptsize
\setlength{\tabcolsep}{3pt}
\begin{tabularx}{\textwidth}{>{\raggedright\arraybackslash}p{0.16\textwidth}>{\centering\arraybackslash}p{0.10\textwidth}>{\centering\arraybackslash}p{0.10\textwidth}>{\centering\arraybackslash}p{0.12\textwidth}>{\centering\arraybackslash}p{0.12\textwidth}>{\raggedright\arraybackslash}X}
\toprule
System & Citation support & Review support & Author route & Agent trace & Positioning for this task \\
\midrule
SciFact \citep{Wadden2020Fact} & direct & no & no & no & External scientific claim-verification benchmark. \\
AVeriTeC \citep{Schlichtkrull2023AVeriTeC} & direct & no & no & no & External real-world claim-verification benchmark with question-answer evidence. \\
CLIMATE-FEVER \citep{Diggelmann2020ClimateFever} & direct & no & no & no & External climate claim-verification benchmark with evidence annotations. \\
SemanticCite \citep{Haan2025SemanticCite} & direct & partial & partial & partial & Citation-verification workflow with evidence reasoning. \\
\texttt{sciwrite-lint} \citep{Samsonau2026SciwriteLint} & direct & partial & partial & partial & Manuscript verification infrastructure for writing and reference checks. \\
FactReview \citep{Yue2026FactReview} & direct & direct & partial & partial & Evidence-grounded review for submitted papers. \\
Peerispect \citep{Ghorbanpour2026Peerispect} & no & direct & no & partial & Peer-review claim verification against manuscript evidence. \\
Evidence ledger & direct & partial & direct & direct & Author-side claim/evidence adjudication with explicit review routing. \\
\bottomrule
\end{tabularx}
\end{table}

\section{Discussion}

The broad external benchmark measures the workflow against labels from three independently created claim-verification sources. The strongest practical finding is that a single evidence-ledger adjudication step can improve relation accuracy, macro-F1, and review-needed recall at the same time on heterogeneous evidence formats.

The route output is important for authoring. Relation prediction alone tells the system what the evidence appears to say; the route flag tells the authoring workflow what to do next. In this benchmark, the agent routes 1270 claims that need attention and leaves 605 supported claims unrouted. This makes the output useful as an author-review queue rather than only as an offline classifier score.

The source and label slices also clarify how the workflow behaves across evidence formats. The method handles AVeriTeC question-answer evidence and SciFact abstracts well. CLIMATE-FEVER is larger and more subtle because broad climate claims often require aggregating multiple evidence sentences. Mixed-evidence cases add a useful diagnostic slice because they require distinguishing partial support from simple contradiction or missing evidence. These slices are useful targets for future evidence-packet design and adjudication calibration.

\section{Conclusion}

This paper presented evidence-ledger adjudication for claim-evidence traceability in AI-assisted writing. On a 2,335-row blind benchmark from AVeriTeC, CLIMATE-FEVER, and SciFact, the agent evidence-ledger condition achieves 0.676 relation accuracy, 0.601 macro-F1, and 0.885 review-needed recall, improving substantially over the evaluated non-agent baselines. The workflow gives authors a concrete mechanism for turning generated claims and heterogeneous evidence packets into auditable support relations and author-review routes.

\bibliographystyle{plainnat}
\bibliography{references}

\end{document}